\documentclass[sigconf]{acmart}

\usepackage{algorithm, algorithmic, graphicx, float}

\copyrightyear{2020}
\acmYear{2020}
\setcopyright{acmcopyright}
\acmConference[CIIS 2020]{2020 The 3rd International Conference on Computational Intelligence and Intelligent Systems}{November 13--15, 2020}{Tokyo, Japan}
\acmBooktitle{2020 The 3rd International Conference on Computational Intelligence and Intelligent Systems (CIIS 2020), November 13--15, 2020, Tokyo, Japan}
\acmPrice{15.00}
\acmDOI{10.1145/3440840.3440860}
\acmISBN{978-1-4503-8808-5/20/11}

\begin{document}

%%
%% The "title" command has an optional parameter,
%% allowing the author to define a "short title" to be used in page headers.
\title{Efficient Low-Latency Dynamic Licensing for Deep Neural Network Deployment on Edge Devices}

%%
%% The "author" command and its associated commands are used to define
%% the authors and their affiliations.
%% Of note is the shared affiliation of the first two authors, and the
%% "authornote" and "authornotemark" commands
%% used to denote shared contribution to the research.
\author{Toan Pham Van}
\email{pham.van.toan@sun-asterisk.com}
\affiliation{%
  \institution{R\&D Lab, Sun* Inc.}
  \city{Hanoi}
  \country{Vietnam}
}

\author{Ngoc N. Tran}
\email{tran.ngo.quang.ngoc@sun-asterisk.com}
\affiliation{%
  \institution{R\&D Lab, Sun* Inc.}
  \city{Hanoi}
  \country{Vietnam}
}

\author{Hoang Pham Minh}
\email{pham.minh.hoang@sun-asterisk.com}
\affiliation{%
  \institution{R\&D Lab, Sun* Inc.}
  \city{Hanoi}
  \country{Vietnam}
}

\author{Tam Nguyen Minh}
\email{nguyen.minh.tam-b@sun-asterisk.com}
\affiliation{%
  \institution{R\&D Lab, Sun* Inc.}
  \city{Hanoi}
  \country{Vietnam}
}

\author{Thanh Ta Minh}
\email{thanhtm@mta.edu.vn}
\affiliation{%
  \institution{Le Quy Don Technical University}
  \city{Hanoi}
  \country{Vietnam}
}

%%
%% By default, the full list of authors will be used in the page
%% headers. Often, this list is too long, and will overlap
%% other information printed in the page headers. This command allows
%% the author to define a more concise list
%% of authors' names for this purpose.
\renewcommand{\shortauthors}{Toan et al.}

%%
%% The abstract is a short summary of the work to be presented in the
%% article.
\begin{abstract}
Along with the rapid development in the field of artificial intelligence (AI), especially deep learning, deep neural network (DNN) applications are becoming more and more popular in reality. To be able to withstand the heavy load from mainstream users, deployment techniques are essential in bringing neural network models from research to production. Among the two popular computing topologies for deploying neural network models in production are cloud-computing and edge-computing. Recent advances in communication technologies, along with the great increase in the number of mobile devices, has made edge-computing gradually become an inevitable trend. In this paper, we propose an architecture to solve deploying and processing deep neural networks on edge-devices by leveraging their synergy with the cloud and the access-control mechanisms of the database. Adopting this architecture allows low-latency DNN model updates on devices. At the same time, with only one model deployed, we can easily make different versions of it by setting access permissions on the model weights. This method allows for dynamic model licensing, which benefits commercial applications.
\end{abstract}

%%
%% The code below is generated by the tool at http://dl.acm.org/ccs.cfm.
%% Please copy and paste the code instead of the example below.
%%


%%
%% Keywords. The author(s) should pick words that accurately describe
%% the work being presented. Separate the keywords with commas.

%% A "teaser" image appears between the author and affiliation
%% information and the body of the document, and typically spans the
%% page.

%%
%% This command processes the author and affiliation and title
%% information and builds the first part of the formatted document.
\maketitle

\section{Introduction}
Deep neural networks (DNNs) have become essential in real-world applications, and thus started the quest for efficient DNN deployment architectures. There are two popular computing topologies for this task: cloud computing and edge computing. Cloud computing \cite{mell2011nist}, a technology that has a long history and numerous achievements, processes everything in the cloud, making it easy for system upgrades, source code updates, and flexible scaling. Deploying DNN models to the cloud would also take advantage of the powerful computing power of these systems; and having only one centralized version for all users would simplify weight updates and model performance monitoring. However, cloud computing also has its limitations in terms of waste of system resources, as it cannot make use of the computing power of edge devices. As a result, the overall cost of maintaining a cloud system is very high. According to the prediction of Ericsson, in 2024, 45\% of global internet data will be generated by the Internet-of-Things (IoT) devices \cite{erricson}. Transferring a great amount of data back-and-forth between edge devices and the cloud is intractable as it causes excessive strain to the network infrastructure. Furthermore, with the development of hardware technology, edge-devices have more and more powerful computing capabilities. For that reason, moving from cloud to edge is historically inevitable.

Instead of cloud computing, pushing DNNs to the edge of the network enhances both efficiency and personalization. The AI model is still trained on the cloud to utilize its great computing power. Afterward, the trained model gets compressed to work with more computationally-limited hardware, then deployed to the edge devices. This setting allows model updates to be distributed quickly, simultaneously on all devices, and while trying to save as much resources as possible. Besides, publishing a model with different licenses is also an interesting issue for commercial applications, where a model can be shipped with different performances depending on the user's license. In this paper, we propose a deployment architecture to deploy the DNN model to edge-devices with four advantages as follows: 
\begin{itemize}
  \item \textbf{Efficient}: In our architecture, we use an in-cloud database to store the DNN  model weights with their corresponding layer names and indices. This design choice helps us centrally manage different versions of weights and easily pushes or updates the changes of DNN model. As a result, the deployment and update process becomes more efficient.
  \item \textbf{Low-Latency Update}: Downloading all the weights of the DNN model can take a long time on edge devices. This operation is only needed the first time when we load model weights onto the edge devices. However, for later updates, we might not need to download all the model weights but only the modified ones. Based on such analysis, we propose an algorithm to download only modified weights to achieve low-latency updates.
  \item \textbf{Version Management}: Managing versions is a common task when deploying a DNN model in production. Our method provides a solution to track changes with commit history, update new versions, and rollback to an older version, similar to Version Control Systems.
  \item \textbf{Dynamic Licensing}: This feature is especially meaningful for commercial applications. Our method allows the model owner to create unlimited licenses of their model with different accuracies but while requires only one set of weights saved on the database.
\end{itemize}

The rest of the paper is organized as follows. Section 2 provides a brief survey of related works. Next, in Section 3, we describe our proposed method. Experiment settings, including dataset information and data processing methods, are described in Section 4; while experimental results are detailed in Section 5. Finally, Section~6 presents our conclusions.

\section{Related Works}
\subsection{Edge-based vs Cloud-based AI}
\label{traditional_edge_ai}
\begin{figure}[t]
    \centering
    \includegraphics[width=9cm,height=5cm]{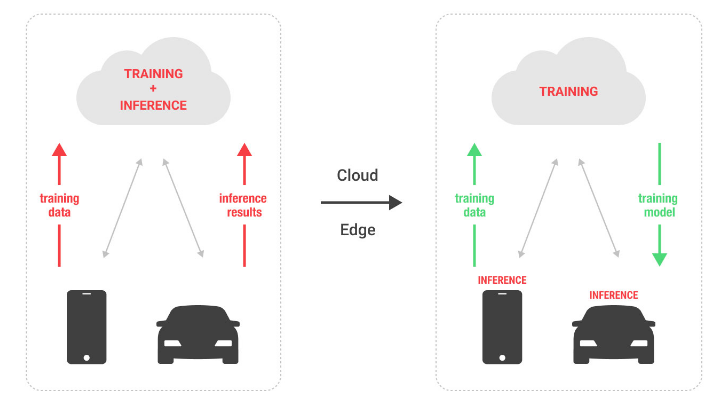}
    \caption{Cloud-based DNN (left) vs Edge-based DNN architecture (right)}
    \label{fig:edge_ai}
\end{figure}

Currently, model inferences are mostly performed in the cloud; but as the diversity of DNN applications grows, other alternatives to the centralized training and inference strategy are required to lessen the burden on network infrastructures \cite{li2018learning}. To that, we opt for Edge Computing \cite{shi2016promise}: a distributed computing paradigm where software-defined networks are built to decentralize data and provide results expected to be the same as which of cloud computing \cite{shi2020communication}. Solving these problems, edge DNN aims to process DNN models directly on edge devices. However, to utilize their innate computing power, edge computing faces more resource allocation problems, due to the inherent difference in hardware architecture, the need to sample inputs from built-in peripherals, bandwidth constraints and more \cite{shi2016edge, shi2020communication}. To offset these problems, deployment on edge-devices requires DNN models to be minimal enough for fast inference and low-latency updates. As a result, it requires various powerful optimizations to achieve the required system efficiency. The difference between cloud-based and edge-based for DNN application is demonstrated in \autoref{fig:edge_ai}. The left figure shows the cloud-based deployment with training, and the right figure shows the edge-based architecture where inference has been offloaded to the edge devices.

\subsection{Deep Neural Network (DNN)}

A deep neural network is, in essence, a composition of more than two non-linear, simpler functions, which can approximate any arbitrarily complex function \cite{10.5555/3086952} (Universal Approximate Theorem \cite{Csji2001ApproximationWA}). Mathematically, a DNN is defined as:
\begin{equation}
f_{\theta(x)}=f_{\theta_{n}} \circ f_{\theta_{2}} \circ f_{\theta_{1}}(x),
\end{equation}
where  $$\theta=\left\{\theta_{1}, \theta_{2}, \ldots, \theta_{n}\right\}$$
is its parameterization.

The most common task is classification, where the model would output whether a data is of a particular class. To build a DNN for the job, we first choose a model architecture that can find meaningful patterns to output. Subsequently, all that is left to do is to learn the parameters of that network through training.

To train a DNN model, a scalar objective function must be defined to quantify the quality of the model; and thus optimizing this function will improve the model’s performance as a result. In the case of classification, we would minimize MLE as our loss function. Optimizing this loss function using gradient-based methods \cite{zhang2019gradient} is a common practice in deep learning algorithms. This ``allows the information from the cost to flow backward through the network to compute the gradient'' \cite{10.5555/3086952}. The optimal parameters (weights) are approximated via the iteration method.

\subsection{Model Compression Methods}

Modern DNNs are expensive in both computation and memory storage, which leads to difficulties in deployment on mobile devices. Consequently, model compression – compressing large, complex models into a lighter, simpler one without significant loss in accuracy – has become indispensable to the DNNs deployment.

\subsubsection{Model Pruning}

An enormously trained model in deep learning contains a large amount of redundancy \cite{cheng2017survey} in the form of unimportant weights that have little contribution to the final output. Pruning is a method of model compression, lightening the architecture by cutting off those unimportant connections or weights, trading a minor loss in quality for performance \cite{cheng2017survey}. Methods of identifying uninformative weights are varied: from using second-order derivatives information \cite{10.5555/109230.109298}, similarity in neurons \cite{srinivas2015datafree}, to the state-of-the-art Movement Pruning \cite{sanh2020movement} which use the change's magnitude in weight updating. Among these above complex methods, one stands out with its simplicity and effectiveness: Magnitude Pruning \cite{Han2016DeepCC} – eliminating weights which have magnitudes less than some threshold, followed by fine-tuning to achieve the highest possible accuracy on the (significantly) pruned model.

\subsubsection{Quantization}

Quantization is the method of reducing the representation accuracy of the network's weights by storing them with lower numbers of bits. This leads to faster computation, decreases in-memory storage with a trade-off of lower accuracy. 8-bit weight quantization \cite{37631} or 16-bit fixed-point with stochastic rounding representation \cite{gupta2015deep} are well-known examples of this method, for their effectiveness heavily outweighs their loss of accuracy \cite{cheng2017survey}. 

\subsubsection{Weight Sharing}
Deep Compression \cite{Han2016DeepCC} also introduces weight-sharing that works well with quantization. This method can be applied after quantization. It divides the weights into $k$ clusters, where weights in one cluster are closed in magnitude, which afterward will be set to be identical. Consequently, a reduction in storage and computation occurs since one only needs to store a sparse cluster-index matrix and a hashtable for quantized value-lookup instead of a dense matrix.

\subsection{Database and Query}
\subsubsection{Database}

A database is a collection of data, typically describing the activities of and among related entities \cite{ramakrishnan2000database}. For different problem requirements and types of information stored, various types of databases are available for the task: relational database, NoSQL database, graph database, etc. \cite{types_of_db}. The most mature and widely used database systems in production today are relational database management systems, which can be found in most applications, such as e-commerce, social networks, retails, etc.

DNNs consist of neurons organized into layers of neurons and the connections between them. This hierarchical architecture encourages us to represent a DNN with a relational database. Storing a DNN in the database would make it easy to update weights, ensuring weight constraints, and manage weight access permissions.

\subsubsection{GraphQL}
Querying from a database fast and efficiently is impossible with the traditional RESTful API \cite{gao2011restful}, so we opted for a better choice for the job: GraphQL \cite{taelman2018graphql}. For all its popularity's worth, one aspect where REST API falls short is its strict inflexible specifications, where custom requests for various information would require a lot of queries to different API endpoints. Not only this would require a complicated control flow to support many types of requests, but also combined latency of all these queries introduces a noticeable lag. Moreover, if one endpoint were to fail, the request would not be able to go through later endpoints – this bottlenecking phenomenon is common with REST API. However, these problems do not occur with GraphQL, being a client-oriented query language with a flexible data structure that can adapt to any type of client's demands. GraphQL supports operations similar to REST API, but with a hierarchical structure that is client-friendly.

\section{Proposed Method}
\subsection{Our Architecture}
\subsubsection{Combining Edge-Devices with weights storage database}
\begin{figure}[b]
    \centering
    \includegraphics[width=7cm,height=7cm]{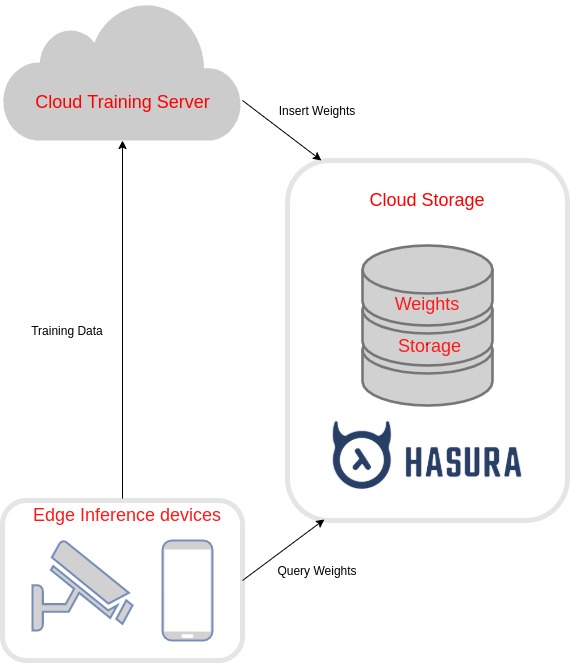}
    \caption{Our architecture with weight storage in database}
    \label{fig:edge_ai_proposed}
\end{figure}

The traditional Edge-AI architecture (mentioned in \autoref{traditional_edge_ai}) only has two main partitions: edge devices and the cloud. The cloud in this computing topology has the roles of training and storing DNN models, managing model versions, storing user data, and model licenses. That is a huge amount of tasks to manage; especially with the model licensing, where the server needs to store a large number of model versions. Imagine we have 10 versions of the model, each of which has 10 different licenses: the cloud would have to store 100 copies of the model to accommodate that. This leads to a waste of system resources. 
\label{server_split}
Therefore, we propose our new pipeline as demonstrated in \autoref{fig:edge_ai_proposed}. Our proposed architecture as the same as an additional module of traditional edge computing in deployment the DNN models. Instead of store all weight and DNN topology in a server, we store the model weights in a database placed on a different cloud. By splitting the traditional single unified cloud into two with their dedicated tasks, we both lessen the workload and let each server do one task and do it well. Moreover, we only store incremental changes across model versions and only license-specific indices of the production model version through our permission management system. Our version management is further elaborated in \autoref{weight_vcs}.

\subsubsection{Updating Weight Versions}
Our system enables a flexible database weight update when there is a newer version available. Our procedure is as follows: we check the existence of each weight variable of the new version and whether its value changed. If a parameter is not yet in the table or its value is different from the old one, a new entry in the \textit{Weight} table will be created to store it. Otherwise, no update in the database will be made. This simple strategy allows significantly more time and storage efficiency in weight-updating, compared to the traditional method of storing parameters of each version in a separate table.

In the other direction, when the database has a newer version than which on the edge device, the user may want to update his version to the newest one (possibly as a response to a push notification). In this situation, the device starts the update by sending our storage server the current model version it has. Then, the server responds with values and indices of the weights that are either newly created or updated. This process is very similar to the previous one, but the benefits of not fetching unchanged weights to reduce extra bandwidth are more noticeable: this be analyzed in more detail later.

\subsection{Model Compression Strategy}
To deploy machine learning models to mobile devices, we lessen the burden of computation and storage by compressing our models before storing them in the database.

\begin{figure*}[ht]
    \centering
    \includegraphics[width=\linewidth]{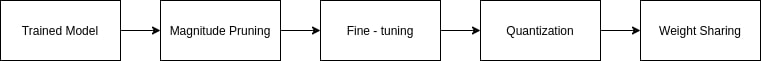}
    \caption{Model compression pipeline}
    \label{fig:compress}
\end{figure*}

\autoref{fig:compress} describes the pipeline of our model compression:
Firstly, we apply magnitude pruning on our models to cut off uninformative weights. Subsequently, we fine-tune models to obtain optimal weights on the pruned architecture. Next, quantization (converting weights from 64-bit to 8-bit representation) followed by weight sharing is used to further compress the models. Despite its simplicity, this pipeline results in highly effective model compression with acceptable performance loss.

\subsection{Weight Database Storage}
\begin{figure*}[t]
    \centering
    \includegraphics[width=\linewidth]{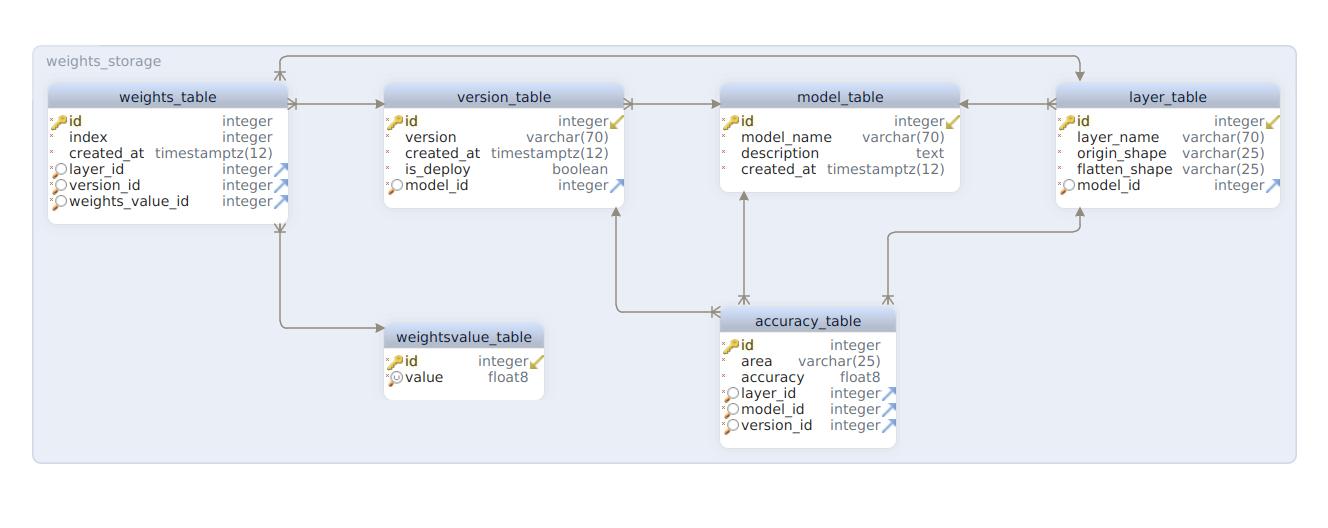}
    \caption{Weight Database Schema}
    \label{fig:schema}
\end{figure*}

After compressing models using our compression pipeline, we store models' weights into Postgres — our database of choice. In order to connect to the Postgres database, we opt for GraphQL API instead of REST API due to its advanced features that allow fast and precise data access. Particularly, Hasura Engine, a GraphQL server, is utilized connect to Postgres in realtime. We use Hasura for querying data from the database, and Django for adding data to the database. 
To harmonize with GraphQL syntaxes, we design our model weight database with tables for \textit{Model}, \textit{Layer}, \textit{Weights}, \textit{Version}, and \textit{Accuracy}, as illustrated in \autoref{fig:schema}.
 
After reducing the model's size with compression, we save the optimized weights in our database along with their metadata: the weight's flattened index, weight value, date added, corresponding \textit{layer} and \textit{version} table entries' foreign keys. Weight arrays extracted from the model instance with common deep learning frameworks are in the same format, being a dictionary with layer names as keys and weight arrays as values, greatly streamlining our storage procedure. After flattening the weight arrays, we save the weights in the aforementioned format. Thanks to model pruning, we only need to save the nonzero weights of the resulting sparse matrices, greatly saving our database's space. This process is time-efficient thanks to Django Object-Relational Mapper's (ORM) bulk insert mechanism.

Conversely, to load weight values from the database, we first build an empty (zeroed) model with its architecture layer-by-layer with their corresponding \textit{Layer}'s table entries. Then, we load the weight values and place them in the layers individually in the locations indicated by their flattened indices.

\subsection{Weight Version Control}
\label{weight_vcs}
During the design process, we might come up with with various versions for one model. For example, during model selection, one may try cross-validation to choose over a range of possible hyperparameters. Changing a hyperparameter set would change every weight in that model, and the new version's record in the database would be a completely new entry. Another possibility is fine-tuning models while freezing layers to incorporate minor improvements: in this case only a subset of weights would change. In this case, the new version only stores the indices and values of the changed weights.

Model versions and their parameters are stored in the \textit{Version} and \textit{Weight} tables respectively. In the \textit{Version} database, there is a status field of the Boolean type which determines whether the major version is used for production. These major versions are completely different from each other, so only one is in use at a time. Then, for minor versions of the production model, each parameter component is linked with its corresponding version ID for efficient weight updates. Specifically, when a model gets an incremental update, only the changed weights are stored in new entries, while the others remain unchanged in the \textit{Weight} database. This improves model retrieving since updating a new version only requires a fraction of the weights to be retrieved.

\textit{Accuracy} table is connected with the \textit{Version}, \textit{Layer}, \textit{Model} tables, and most importantly, the \textit{Weight} table. In \textit{Accuracy}, for a particular version accuracy value, we save a set of weight value ranges to be used that would give us the desired performance. The idea is that we artificially cause worse model performance by not using weights in some particular magnitude ranges. With information from the requesting user's license, a version with the corresponding accuracy will be shipped.

\subsection{Weight Licensing}
Given that machine learning models are our product, we need to have a customer licensing mechanism for proprietary reasons. A tangible example would be as follows: we offer a 3-layer-perceptron model with 98\% accuracy through a freemium business model. Using a permission control mechanism, we do not let free-tier users to access weights of the first layers with magnitudes between $0.5$ and $0.8$; instead, these weights would be set to 0 (similar to the pruning process). The model's accuracy then drops to 70\% in our experiment.

The aforementioned \textit{Accuracy} table contains masks over our deployed machine learning model's weights. The more weights the mask hides, the worst accuracy a version has.
To get weight indices corresponding to some specific desired accuracy, all current deployed model's weight values are divided into equal-sized intervals based on their magnitude. We perform gradual magnitude pruning on the model, results in a corresponding gradual accuracy reduction. The pruning process terminates when the required accuracy is observed. Algorithm \ref{algo:prune} describes our method of retrieving pruned models based on desired accuracy.

\begin{algorithm}[ht]
\begin{algorithmic}
    \STATE divide weight range into $k$ smaller equal-sized intervals
    \STATE initialize a list of cut-off intervals
    \FORALL{intervals} %\COMMENT{gradually cut off unimportant weights of model}
        \FORALL{model's layers}
            \STATE{cut off weights that have values in that interval}
        \STATE append interval into cut-off interval list
            \IF{accuracy of pruned model is close to the target}
            	\STATE break the pruning process
            \ENDIF
        \ENDFOR
    \ENDFOR
    \RETURN{uncut interval lists}
\end{algorithmic}
\caption{Pruning model based on accuracy}
\label{algo:prune}
\end{algorithm}

For customers with licenses that fall into our predefined tiers, we use \textbf{Static Licensing}: we access the \textit{Accuracy} table, which is directly linked to the \textit{Weights} table and stores the weight ranges for every layer and their corresponding accuracies. These weight range-accuracy relationships are evaluated beforehand; and at deployment only lookups are needed to ship the appropriate set of weights according to the user's subscription tier. On the other hand, if the client requires custom performance tier, we turn to \textbf{Dynamic Licensing}, which evaluates these ranges on-demand.

\section{Main Features of Our Method}
\subsection{Efficient Deployment}
\label{efficient_deploy}
As mentioned in \autoref{server_split}, we break the work that is traditionally all processed in one cloud server into two highly coherent tasks. Even with decent scaling solutions like Docker Swarm or Kubernetes, if the basic worker instance has to work with both training and shipping new weights, it will be a lot less efficient than having separate instances doing their respective tasks.

\subsection{Version Management}
Our design stores weight incrementally, instead of a completely new whole model every new version. With this, not only we make the previously analyzed \nameref{efficient_deploy} possible, we save storage by only keeping changes and not the unchanged weights. Moreover, this allows for skipping intermediate patches. Given that every weight entry in our database is stored along with the version it was last updated, the customer can query for all new weights throughout his missed updates in one go, instead of having to download individual version updates and apply them one-by-one gradually.

\subsection{Low-Latency Update}
This approach's contribution to fast evaluation is manifold: first, the customer's model evaluation is carried out within their device. This is crucial for time-critical operations, such as self-driving cars, where milliseconds matter. This is also important to always-on systems, such as health gadgets with real-time illness detection -- a model evaluation server experiencing difficulty at the worst possible time may cost lives. Moreover, the data for evaluation stays on the customer's device, which is a huge plus for the privacy-conscious. For us, this also offloads work to the client's device, which is getting more and more capable as technology develops. Second, our setup allows fast weight updates from our database server. This would greatly benefit interdisciplinarity that value timely updates of even minuscule incremental performance, such as stock trading. For large hedge funds, even just a 0.1\% percent increase in accuracy shipped 1 millisecond earlier could mean millions of dollars.

\subsection{Dynamic and Static Licensing}
Most businesses offer their service tiers in terms of speed and/or availability, but here we introduce a new factor: accuracy. Most trial users would want a not-too-accurate but fast free-tier service since their applications would not need more than that, while their patience is limited. Moreover, our design allows dynamic licensing which lets the clients custom their version's performance beyond our predefined tiers. This allows flexibility to match whatever the most budget-conscious client needs.

\section{Experiments}

\subsection{Experimental Settings}
For our experiments, we use the following technologies:
\begin{itemize}
\item Django Framework: database connecter for inserting and updating data into database with Django ORM \cite{forcier2008python}.
\item Keras \cite{keras}, TensorFlow \cite{tensorflow}: deep learning frameworks for building DNN architectures.
\item PostgreSQL \cite{postgres}: a relational database backend,
\item Hasura \cite{hasura}: an instant GraphQL engine with built-in authorization for querying weights,
\item Docker \cite{docker}: machine virtualization for simulating real case studies in this paper.
\end{itemize}

To verify our system's effectiveness, we measure the amount of space needed to store weights in our database with gradual additions of various optimizations. The results of these experiments are listed in \autoref{tab:storage}.

\subsection{Experimental Results}

\begin{table}[h]
\caption{The cost of memory storage}
\label{tab:storage} \centering
\tabcolsep3pt\begin{tabular}{cccp{60pt}}
  \toprule
  \textbf{No. of params} & \textbf{Full params} & \textbf{Pruning 80\%} & \textbf{Pruning 80\% +\hfill\break Quantization} \\
  \midrule
  109386 & 13MB & 2.92MB & 2.34MB\\
  \midrule
  101770 & 12MB & 2.65MB & 2.09MB\\
  \bottomrule
\end{tabular}
\end{table}

\autoref{tab:storage} shows our results in saving weights of models with various optimizations. With more than 100,000 weights, saving all weights in Postgres takes 13MB of space. Through pruning the unimportant 80\% of the weights and not storing their unused bits with a more concise representation, the storage size reduces to 2.09MB.

\section{Conclusion}
In this paper, we successfully design an architecture utilizing an In-cloud database for efficient deployment and centralization management. The merits of our procedure are additionally represented by low-latency update (which only downloads modified weights on to edge devices), flexible version managed (allows track on history commitment changes, new version update, old version rollback), dynamic and static licensing (provides both ready-to-use and custom-accuracy models). With this novel deployment architecture, we hope our contributions will be beneficial to numerous DNN applications. 

%%\section*{Acknowledgment}
\begin{acks}
This work is partially supported by \textbf{Sun-Asterisk Inc}. We would like to thank our colleagues at \textbf{Sun-Asterisk Inc} for their advice and expertise. Without their support, this experiment would not have been accomplished.
\end{acks}
%%
%% The next two lines define the bibliography style to be used, and
%% the bibliography file.
\bibliographystyle{ACM-Reference-Format}
\bibliography{references}
% \printbibliography

\end{document}